\documentclass[11pt,a4paper]{article}
\usepackage[hyperref]{ranlp2021}
\usepackage{times}
\usepackage{latexsym}

\usepackage{graphicx}
\usepackage{amsmath}
\usepackage{nccmath}
\usepackage{amssymb}
\usepackage{mathtools}
\usepackage{multirow}
\usepackage{xcolor}

\usepackage{microtype}

\aclfinalcopy 

\title{Improving Distantly Supervised Relation Extraction\\ with Self-Ensemble Noise Filtering}

\author{Tapas Nayak \\
  IIT Kharagpur \\
  India \\\And
  Navonil Majumder \\
  SUTD \\
  Singapore \\
  \texttt{\{tnk02.05,n.majumder.2009,soujanya.poria\}@gmail.com} \\\And
  Soujanya Poria \\ 
  SUTD \\
  Singapore}

\date{}

\begin{document}
\maketitle
\begin{abstract}

Distantly supervised models are very popular for relation extraction since we can obtain a large amount of training data using the distant supervision method without human annotation. In distant supervision, a sentence is considered as a source of a tuple if the sentence contains both entities of the tuple. However, this condition is too permissive and does not guarantee the presence of relevant relation-specific information in the sentence. As such, distantly supervised training data contains much noise which adversely affects the performance of the models. In this paper, we propose a self-ensemble filtering mechanism to filter out the noisy samples during the training process. We evaluate our proposed framework on the New York Times dataset which is obtained via distant supervision. Our experiments with multiple state-of-the-art neural relation extraction models show that our proposed filtering mechanism improves the robustness of the models and increases their F1 scores.

\end{abstract}

\section{Introduction}

The task of relation extraction is about finding relation or no relation between two entities. This is an important task to fill the gaps of existing knowledge bases (KB). Open information extraction (OpenIE) \cite{banko2007open} is one way of extracting relations from text. They consider the verb in a sentence as the relation and then find the noun phrases located to the left and right of that verb as the entities. But this process has two serious problems: First, the same relation can appear in the text with many verb forms and OpenIE treats them as different relations. This leads to the duplication of relations in KB. Second, OpenIE treats any verbs in a sentence as a relation which can generate a large number of insignificant tuples which cannot be added to a KB. 

Supervised relation extraction models, on the other hand, do not have these problems. But they require a large amount of annotated data which is difficult to get. \newcite{mintz2009distant}, \newcite{riedel2010modeling}, and \newcite{hoffmann2011knowledge} used the idea of distant supervision to automatically obtain the training data to overcome this problem. The idea of distant supervision is that if a sentence contains both the entities of a tuple, it is chosen as a source sentence of this tuple. Although this process can generate some noisy training instances, it can give a significant amount of training data which can be used to build supervised models for this task. They map the tuples from existing KBs such as Freebase \cite{bollacker2008freebase} to the text corpus such as Wikipedia articles \cite{mintz2009distant} or New York Times articles \cite{riedel2010modeling,hoffmann2011knowledge}. 

Based on distantly supervised training data, researchers have proposed many state-of-the-art models for relation extraction. \newcite{mintz2009distant}, \newcite{riedel2010modeling}, and \newcite{hoffmann2011knowledge} proposed feature-based learning models and used entity tokens and their nearby tokens, their part-of-speech tags, and other linguistic features to train their models. Recently, many neural network-based models have been proposed to avoid feature engineering. \newcite{zeng2014relation} and \newcite{zeng2015distant} used convolutional neural networks (CNN) with max-pooling to find the relation between two given entities. \newcite{huang2016attention}, \newcite{jat2018attention}, \newcite{nayak2019effective} used attention framework in their neural models for this task. 

\begin{table*}[ht]
\small
\centering
\begin{tabular}{l|c|c|c|c|c}
\hline
\multicolumn{1}{c|}{Sentence}                                                                                           & Entity 1                                                & Entity 2                                               & \begin{tabular}[c]{@{}c@{}}DS Relation\end{tabular} & \begin{tabular}[c]{@{}c@{}}Actual\\ Relation\end{tabular} & Status \\ \hline
\begin{tabular}[c]{@{}l@{}}\textcolor{red}{Barack Obama} was born \\in \textcolor{blue}{Hawaii} .\end{tabular}                                             & \textcolor{red}{Barack Obama}  & \textcolor{blue}{Hawaii}                                                 & birth\_place                                                          & birth\_place                                          & Clean         \\ \hline
\begin{tabular}[c]{@{}l@{}}\textcolor{red}{Barack Obama} visited \\\textcolor{blue}{Hawaii} .\end{tabular}                                                 & \textcolor{red}{Barack Obama} & \textcolor{blue}{Hawaii}                                                 & birth\_place                                                          & None                                                      & Noisy        \\ \hline
\begin{tabular}[c]{@{}l@{}}Suvendu Adhikari was \\born at \textcolor{red}{Karkuli} in Purba \\Medinipur in \textcolor{blue}{West Bengal} .\end{tabular} & \textcolor{red}{Karkuli}                                                 & \textcolor{blue}{West Bengal} & None                                                                      & located\_in                                           & Noisy        \\ \hline
\begin{tabular}[c]{@{}l@{}}Suvendu Adhikari, transport \\minister of \textcolor{blue}{West Bengal}, \\visited \textcolor{red}{Karkuli} .\end{tabular}  & \textcolor{red}{Karkuli}                                                 & \textcolor{blue}{West Bengal} & None                                                                      & None                                                      & Clean         \\ \hline
\end{tabular}
\caption{Examples of distantly supervised (DS) clean and noisy samples.}
\label{tab:noisy_examples}
\end{table*}

But the distantly supervised data may contain many noisy samples. Sometimes sentences may contain the two entities of a positive tuple, but they may not contain the relation specific information. These kinds of sentences and entity pairs are considered as positive noisy samples. Another set of noisy samples comes from the way samples for {\em None} relation are created. If a sentence contains two entities from the KB and there is no positive relation between these two entities in the KB, this sentence and entity pair is considered as a sample for {\em None} relation. But knowledge bases are not complete and many valid relations among the entities in the KBs are missing. So it may be possible that the sentence contains information about some positive relation between the two entities, but since the relation is not present in the KB, this sentence and entity pair is incorrectly considered as a sample for {\em None} relation. These kinds of sentences and entity pairs are considered as negative noisy samples. 

We include examples of clean and noisy samples generated using distant supervision in Table \ref{tab:noisy_examples}. The KB contains many entities out of which four entities are {\em Barack Obama}, {\em Hawaii}, {\em Karkuli}, and {\em West Bengal}. {\em Barack Obama} and {\em Hawaii} have a {\em birth\_place} relation between them. Karkuli and West Bengal are not connected with any relations in the KB. So we assume that there is no valid relation between these two entities. The sentence in the first sample contains the two entities {\em Barack Obama} and {\em Hawaii}, and it also contains information about {\em Obama} being born in {\em Hawaii}. So this sentence is a correct source for the tuple ({\em Barack Obama}, {\em Hawaii}, {\em birth\_place}). So this is a positive clean sample. The sentence in the second sample contains the two entities, but it does not contain the information about {\em Barack Obama} being born in {\em Hawaii}. So it is a positive noisy sample. In the case of the third and fourth samples, according to distant supervision, they are considered as samples for {\em None} relation. But the sentence in the third sample contains the information for the relation {\em located\_in} between {\em Karkuli} and {\em West Bengal}. So the third sample is a negative noisy sample. The fourth sample is an example of a negative clean sample.

The presence of the noisy samples in the distantly supervised data adversely affects the performance of the models. Our goal is to remove the noisy samples from the training process to make the models more robust for this task. We propose a self-ensemble based noisy samples filtering method for this purpose. Our framework identifies the noisy samples during the training and removes them from training data in the following iterations. This framework can be used with any supervised relation extraction model. We run experiments with several state-of-the-art neural models, namely Convolutional Neural Network (CNN) \cite{zeng2014relation}, Piecewise Convolutional Neural Network (PCNN) \cite{zeng2015distant}, Entity Attention (EA) \cite{huang2016attention}, and Bi-GRU Word Attention (BGWA) \cite{jat2018attention} with the distantly supervised New York Times dataset \cite{hoffmann2011knowledge}. Our framework improves the F1 score of these models by $2.1\%$, $1.1\%$, $2.1\%$, and $2.3\%$ respectively\footnote{The code and data for this work is available at https://github.com/nayakt/SENF4DSRE.git}.

\section{Task Description}

Sentence-level relation extraction is defined as follows: Given a sentence $S$ and two entities $\{E_1, E_2\}$ marked in the sentence, find the relation $r(E_1,E_2)$ between these two entities in $S$ from a pre-defined set of relations $R \cup \{\mathit{None}\}$. $R$ is the set of positive relations and {\em None} indicates that none of the relations in $R$ holds between the two marked entities in the sentence. The relation between the entities is argument order-specific, i.e., $r(E_1,E_2)$ and $r(E_2,E_1)$ are not the same. The input to the system is a sentence $S$ and two entities $E_1$ and $E_2$, and output is the relation $r(E_1,E_2) \in R \cup \{\mathit{None}\}$. Distant supervised datasets are used for training relation extraction models. But the presence of noisy samples negatively affects their performance. In this work, we try to identify these noisy samples during training and filter them out from the subsequent training process to improve the performance of the models.

\section{Self-Ensemble Filtering Framework}

\begin{figure*}[ht]
\centering
\includegraphics[scale=0.5]{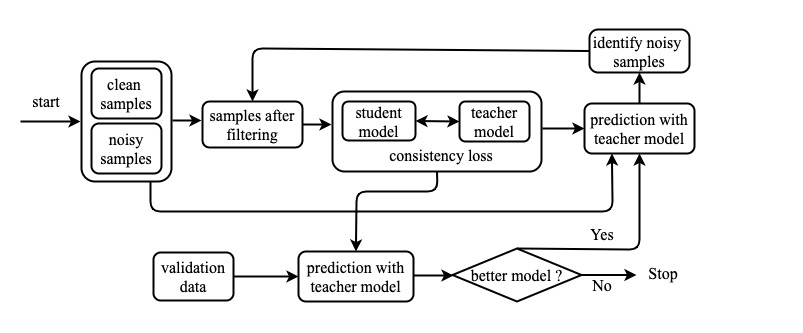}
\caption{Overview of the self-ensemble noisy samples filtering framework. It starts with the clean and noisy samples generated by distant supervision. During training, a self-ensemble version of the model is maintained. At the end of an iteration, this self-ensemble model is used to identify the noisy samples in the training data. These noisy samples are filtered out from the next iteration of training.}
\label{fig:filtering_model}
\end{figure*}

Figure \ref{fig:filtering_model} shows our proposed self-ensemble filtering framework. This framework is inspired from the work by \newcite{Nguyen2020SELF}. We start with clean and noisy samples and assume that all samples are clean. At the end of each iteration, we predict the labels of the entire training samples. Based on the predicted label and the label assigned by distant supervision, we decide to filter out a sample in the next iteration. After each iteration, we consider the entire training samples for the filtering process. The individual models at each iteration can be very sensitive to wrong labels, so in our training process, we maintain a self-ensemble version of the models which is a moving average of the models of previous iterations. We hypothesize that the predictions of the ensemble model are more stable than the individual models. So the predictions from the ensemble model are used to identify the noisy samples. These noisy samples are removed from the training samples of the next iteration. We consider the entire distantly supervised training data for prediction and filtering so that if a sample is filtered out wrongly in an iteration, it can be included again in the training data in the subsequent iteration. 

\subsection{Self-Ensemble Training}

We use the student-teacher training mechanism proposed by \newcite{Tarvainen2017MeanTA} for our self-ensemble model learning. A student model can be any supervised learning model such as a neural network model. A teacher model is the clone of student model with same parameters. The weights of parameters of this teacher model is the exponential moving average of the weights of parameters of the student model. So this teacher model is the self-ensemble version of the student model. An additional consistency loss is used to maintain the consistency of the predictions of the student model and the teacher model. Following is the step-by-step algorithm to train such an self-ensemble model:
\begin{enumerate}
    \item First, a student model $M_s^i$ is initialized. This can be any supervised relation extraction model such as CNN, PCNN, Entity Attention (EA) or Bi-GRU Word Attention (BGWA) model.
    \item A teacher model $M_t^i$ is cloned from the student model $M_s^i$. We completely detach the weights of the teacher model from the student model.
    \item A gradient descent based optimizer is selected to update the parameters of the student model.
    \item Loss is calculated based on the cross-entropy loss of the student model for the classification task and a consistency loss between the student model and teacher model.
    \item In each training iteration or epoch:
        \begin{itemize}
            \item In each step or mini-batch:
                \begin{itemize}
                    \item Update the weights of the student model $M_s^i$ using the selected optimizer and the loss function.
                    \item Update the weights of the teacher model $M_t^i$ as an exponential moving average of the student weights.
                    \end{itemize}{}
            \item Evaluate the performance of the teacher model $M_t^i$ on a validation dataset. If we decide to continue the training after evaluation, we use a filtering strategy at this point to remove the noisy samples from the training data. This clean training data is used in the next iteration of the training process.
        \end{itemize}
    \item Return the best teacher model $M_t^i$. This teacher model is the self-ensemble version of the student model.
\end{enumerate}

\subsection{Loss Function \& Updating the Student}

We use the negative log-likelihood loss of the relation classification task from the student model ($\mathcal{L}_{ce}$ ) and a mean-squared error based consistency loss between the student and teacher model ($\mathcal{L}_{mse}$) to update the student model.
\begin{align*}
&\mathcal{L}_{ce} = -\frac{1}{B} \sum_{i=1}^{B} \mathrm{log} (p(r_{i} \vert s_i, e_i^1, e_i^2, \theta_s))\\
&\mathcal{L}_{mse} = \frac{1}{B} \sum_{i=1}^{B} \sum_{j=1}^{C} ( y_s^{i,j} -y_t^{i,j})^2\\
&\mathcal{L} = \mathcal{L}_{ce} + \mathcal{L}_{mse}
\end{align*}
\noindent For $\mathcal{L}_{ce}$, $p(r_{i} \vert s_i, e_i^1, e_i^2, \theta_s)$ is the conditional probability of the true relation $r_i$ when the sentence $s_i$, two entities $e_i^1$ and $e_i^2$, and the model parameters of the student $\theta_s$ are given. For $\mathcal{L}_{mse}$, $y_s^{i,j}$ and $y_t^{i,j}$ are the softmax output of the $j$ th relation class of $i$ th training sample in the batch from the student model and the teacher model respectively. $C$ is number of relation class in the dataset and $B$ is the batch size. The parameters of the student model $\theta_s$ are updated based on the combined loss $\mathcal{L}$ using an gradient descent based optimizer. The consistency loss ($\mathcal{L}_{mse}$) makes sure that output softmax distribution of the student model and teacher model are close to each other, thus maintain the consistency of the output from both models.

\subsection{Updating the Teacher}

We update the parameters of teacher model $\theta_t$ based on the exponential moving average of the all previous optimization steps of the student model.
\begin{align*}
    \mathbf{W}(\theta_t^l) = \alpha \mathbf{W}(\theta_t^{l-1}) + (1 - \alpha) \mathbf{W}(\theta_s^l)
\end{align*}
\noindent where $\mathbf{W}(\theta_t^l)$ and $\mathbf{W}(\theta_s^l)$ are the weights of the parameters of the teacher model and student model respectively after the $l$ th global optimization step. $\mathbf{W}(\theta_t^{l-1})$ is the weights of the teacher model parameters up to the $l-1$ th global optimization step. $\alpha$ is a weight factor to control the contribution of the student model of the current step and the teacher model up to the previous step. At the initial optimization steps of the training, we keep the value of $\alpha$ low as the self-ensemble model or teacher model is not stable yet and the student model should contribute more. As the training progress and the self-ensemble model becomes stable, we slowly increase the value of $\alpha$ so that we take the majority contribution from the self-ensemble model itself. We use the following Gaussian curve \cite{He2018AdaptiveSL} to ramp up the value of $\alpha$ from 0 to $\alpha_{max}$ which is a hyper-parameter of the model.

\begin{align*}
    &T = E * \lceil{\frac{L}{B}\rceil} \\
    &p = 1-\frac{\text{min}(\text{step\_idx}, T)}{T} \\
    &\alpha = e^{-5p^2} \alpha_{max}
\end{align*}{}

\noindent Here $E$ is the epoch count to ramp up the $\alpha$ from 0 to $\alpha_{max}$. $E$ is a hyper-parameter of the model and generally, this is lower than the total number of epochs of the training process. $L$ is the size of distant supervised training data at the beginning of training, $B$ is the batch size, and $\text{step\_idx}$ is the current global optimization step count of the training. $T$ represents the number of global optimization steps required for $\alpha$ to reach its maximum value $\alpha_{max}$.

\begin{figure}[ht]
\centering
\includegraphics[scale=0.5]{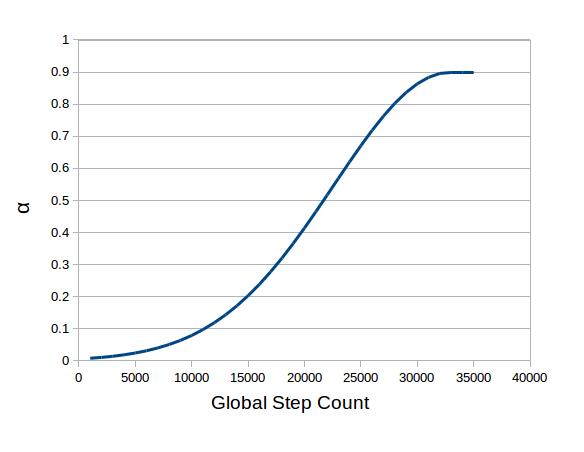}
\caption{Ramping up of $\alpha$ during training. We use E=5, T=33,000, and $\alpha_{max}=0.9$ to generate this curve for the demonstration of how $\alpha$ reaches from 0 to $\alpha_{max}$.}
\label{fig:rampup}
\end{figure}

\subsection{Noise Filtering Strategy}

After each iteration, we use a validation dataset to determine to stop or to continue the training. If we decide to continue the training, then we use the self-ensemble model or the teacher model to filter out noisy samples from the initial training data. This clean training data is used in the next training iteration. We use the self-ensemble model to predict the relation on initial training data for the filtering process after each iteration. We use the entire initial training data for prediction so that if a training sample is filtered out wrongly in an iteration as a noisy one, it can be used again in subsequent training iterations if the subsequent self-ensemble model predicts the sample as a clean one.

Generally, distantly supervised datasets contain a largely high number of {\em None} samples than the valid relation samples. For this reason, we choose a strict filtering strategy for {\em None} samples and a lenient filtering strategy for valid relation samples. We consider a {\em None} sample as clean if teacher models predict the {\em None} relation. Otherwise, this sample is considered as noisy and filtered out from the training set of next iteration. For the valid relations, we consider a sample as clean if the relation assigned by distant supervision belongs to the top $K$ predictions of the teacher model. This clean training data is used in the next training iteration. 

\section{Student Models}

We have used the following state-of-the-art neural relation extraction models as the student model in our filtering framework. These models use three types of embedding vectors: (1) word embedding vector $\mathbf{w} \in \mathbb{R}^{d_w}$ (2) a positional embedding vector $\mathbf{u}^1 \in \mathbb{R}^{d_u}$ which represents the linear distance of a word from the start token of entity $1$ (3) another positional embedding vector $\mathbf{u}^2 \in \mathbb{R}^{d_u}$ which represents the linear distance of a word from the start token of entity $2$. The sentences are represented using a sequence of vectors  $\{\mathbf{x}_1, \mathbf{x}_2,....., \mathbf{x}_n\}$ where $\mathbf{x}_t = \mathbf{w}_t \Vert \mathbf{u}_t^1 \Vert \mathbf{u}_t^2$. $\Vert$ represents the concatenation of vectors and $n$ is the sentence length. These token vectors $\mathbf{x}_t$ are given as input to all the following models.

\subsection{CNN \cite{zeng2014relation}}

In this model, convolution operations with max-pooling are applied on the token vectors sequence $\{\mathbf{x}_1, \mathbf{x}_2,....., \mathbf{x}_n\}$ to obtain the sentence-level feature vector.

\begin{align*}
    &c_i = \mathbf{f}^T (\mathbf{x}_{i} \Vert \mathbf{x}_{i + 1} \Vert .... \Vert \mathbf{x}_{i+k-1}) \\
    &c_{max} = \text{max}(c_1,c_2,....,c_{n}) \\
    &\mathbf{v} = [c_{max}^1, c_{max}^2, ...., c_{max}^{f_k}]
\end{align*} 
$\mathbf{f}$ is a convolutional filter vector of dimension $k(d_w+2d_u)$ where $k$ is the filter width. The index $i$ moves from $1$ to $n$ and produces a set of scalar values $\{c_1, c_2, .....,c_{n}\}$. The max-pooling operation chooses the maximum $c_{max}$ from these values as a feature. With $f_k$ number of filters, we get a feature vector $\mathbf{v} \in \mathbb{R}^{f_k}$. This feature vector $\mathbf{v}$ is passed to feed-forward layer with softmax to classify the relation.

\subsection{PCNN \cite{zeng2015distant}}    

Piecewise Convolutional Neural Network (PCNN) is a modified version of the CNN model described above. Similar to the CNN model, convolutional operations are applied to the input vector sequence. But CNN and PCNN models differ on how the max-pooling operation is performed on the convolutional outputs. Rather than applying a global max-pooling operation on the entire sentence, three max-pooling operations are applied on three segments/pieces of the sentence based on the location of the two entities. This is why this model is called the Piecewise Convolutional Neural Network (PCNN). The first max-pooling operation is applied from the beginning of the sequence to the end of the entity appearing first in the sentence. The second max-pooling operation is applied from the beginning of the entity appearing first in the sentence to the end of the entity appearing second in the sentence. The third max-pooling operation is applied from the beginning of the entity appearing second in the sentence to the end of the sentence. These max-pooled features are concatenated and passed to a feed-forward layer with softmax to determine the relation. 
    
\subsection{Entity Attention (EA) \cite{huang2016attention}}

This model combines the CNN model with an attention network. First, convolutional operations with max-pooling are used to extract the global features of the sentence. Next, attention is applied to the words of the sentence based on the two entities separately. The word embedding of the last token of an entity is concatenated with the embedding of every word. This concatenated representation is passed to a feed-forward layer with tanh activation and then another feed-forward layer with softmax to get a scalar attention score for every word for that entity. The word embeddings are averaged based on the attention scores to get the attentive feature vectors. The CNN-extracted global feature vector and two attentive feature vectors for the two entities are concatenated and passed to a feed-forward layer with softmax to determine the relation.

\subsection{Bi-GRU Word Attention (BGWA) \cite{jat2018attention}}

This model uses a bidirectional gated recurrent unit (Bi-GRU) \cite{cho2014properties} to capture the long-term dependency among the words in the sentence. The tokens vectors $\mathbf{x}_t$ are passed to a Bi-GRU layer. The hidden vectors of the Bi-GRU layer are passed to a bi-linear operator which is a combination of two feed-forward layers with softmax to compute a scalar attention score for each word. The hidden vectors of the Bi-GRU layer are multiplied by their corresponding attention scores for scaling up the hidden vectors. A piecewise convolution neural network \cite{zeng2015distant} is used on top of the scaled hidden vectors to obtain the feature vector. This feature vector is passed to a feed-forward layer with softmax to determine the relation.

\section{Experiments}

\subsection{Datasets}

To verify our hypothesis, we need training data that is created using distant supervision, thus noisy and test data which is not noisy, thus human-annotated. If the test data is also noisy, then it will be hard to derive any conclusion from the results. So, we choose the New York Times (NYT) corpus of \newcite{hoffmann2011knowledge} for our experiments. This dataset has $24$ valid relations and a {\em None} relation. The statistics of the dataset is given in Table \ref{tab:dataset}. The training dataset is created by aligning Freebase tuples to NYT articles, but the test dataset is manually annotated. We use $10\%$ of the training data as validation data and the remaining $90\%$ for training.

\begin{table}[ht]
\small
\centering
\begin{tabular}{l|cc}
\hline
                        & Train   & Test  \\ \hline
\#valid relations       & 24      & 24    \\ 
\#valid relation instances & 100,671 & 520   \\ 
\#None relation instances  & 235,172 & 930   \\ \hline
\end{tabular}
\caption{The statistics of the NYT dataset.}
\label{tab:dataset}
\end{table}

\subsection{Evaluation Metrics}

We use precision, recall, and F1 scores to evaluate the performance of models on relation extraction after removing the {\em None} labels. We use a confidence threshold to decide if the relation of a test instance belongs to the set of valid relations $R$ or {\em None}. If the network predicts {\em None} for a test instance, then it is considered as {\em None} only. But if the network predicts a relation from the set $R$ and the corresponding softmax score is below the confidence threshold, then the final predicted label is changed to {\em None}. This confidence threshold is the one that achieves the highest F1 score on the validation data. 

\begin{table*}[ht]
\small
\centering
\begin{tabular}{l|ccc|cccc}
\hline
      & \multicolumn{3}{c|}{Student}                                                                                                                                                      & \multicolumn{4}{c}{SEF}                                                                                                                                                                       \\ \hline
Model & Prec.                                                     & Rec.                                                      & F1                                                        & Prec.                                                     & Rec.                                                      & F1                                                        & $\uparrow$ \\ \hline
CNN   & \begin{tabular}[c]{@{}c@{}}0.451\\ $\pm$\\ 0.015\end{tabular} & \begin{tabular}[c]{@{}c@{}}0.607\\ $\pm$\\ 0.033\end{tabular} & \begin{tabular}[c]{@{}c@{}}0.518\\ $\pm$\\ 0.021\end{tabular} & \begin{tabular}[c]{@{}c@{}}0.452\\ $\pm$\\ 0.011\end{tabular} & \begin{tabular}[c]{@{}c@{}}0.669\\ $\pm$\\ 0.016\end{tabular} & \begin{tabular}[c]{@{}c@{}}0.539\\ $\pm$\\ 0.005\end{tabular} & 2.1\%      \\ \hline
PCNN  & \begin{tabular}[c]{@{}c@{}}0.431\\ $\pm$\\ 0.013\end{tabular} & \begin{tabular}[c]{@{}c@{}}0.673\\ $\pm$\\ 0.007\end{tabular} & \begin{tabular}[c]{@{}c@{}}0.526\\ $\pm$\\ 0.010\end{tabular} & \begin{tabular}[c]{@{}c@{}}0.432\\ $\pm$\\ 0.009\end{tabular} & \begin{tabular}[c]{@{}c@{}}0.708\\ $\pm$\\ 0.016\end{tabular} & \begin{tabular}[c]{@{}c@{}}0.537\\ $\pm$\\ 0.011\end{tabular} & 1.1\%      \\ \hline
EA    & \begin{tabular}[c]{@{}c@{}}0.437\\ $\pm$\\ 0.012\end{tabular} & \begin{tabular}[c]{@{}c@{}}0.653\\ $\pm$\\ 0.016\end{tabular} & \begin{tabular}[c]{@{}c@{}}0.523\\ $\pm$\\ 0.008\end{tabular} & \begin{tabular}[c]{@{}c@{}}0.444\\ $\pm$\\ 0.008\end{tabular} & \begin{tabular}[c]{@{}c@{}}0.702\\ $\pm$\\ 0.014\end{tabular} & \begin{tabular}[c]{@{}c@{}}0.544\\ $\pm$\\ 0.009\end{tabular} & 2.1\%      \\ \hline
BGWA  & \begin{tabular}[c]{@{}c@{}}0.414\\ $\pm$\\ 0.006\end{tabular} & \begin{tabular}[c]{@{}c@{}}0.680\\ $\pm$\\ 0.021\end{tabular} & \begin{tabular}[c]{@{}c@{}}0.515\\ $\pm$\\ 0.010\end{tabular} & \begin{tabular}[c]{@{}c@{}}0.430\\ $\pm$\\ 0.005\end{tabular} & \begin{tabular}[c]{@{}c@{}}0.720\\ $\pm$\\ 0.014\end{tabular} & \begin{tabular}[c]{@{}c@{}}0.538\\ $\pm$\\ 0.007\end{tabular} & 2.3\%      \\ \hline
\end{tabular}
\caption{Precision, Recall, and F1 score comparison of the student models on NYT dataset when trained with self-ensemble filtering framework (SEF column) and when trained independently (Student column). We report the average of five runs with standard deviation. $\uparrow$ column shows the absolute \% improvement of F1 score over the Student models.}
\label{tab:results}
\end{table*}

\subsection{Parameter Settings}

We run word2vec \cite{mikolov2013efficient} on the NYT corpus to obtain the initial word embeddings with a dimension of $d_w=50$ and update the embeddings during training. We set the dimension positional embedding vector at $d_u=5$. We use $f_k=230$ convolutional filters of kernel size $k=3$ for feature extraction whenever we apply the convolution operation. We use dropout in our network with a dropout rate of $0.5$, and in convolutional layers, we use the tanh activation function. We train our models with a mini-batch size of $50$ and optimize the network parameters using the Adagrad optimizer \cite{duchi2011adaptive}. We want to keep the value of $\alpha_{max}$ high because when the training progress, we want to increase the contribution of the self-ensemble model compare to the student model. So we set the value of $\alpha_{max}$ at 0.9. We experiment with $E=\{5,10\}$ epochs to ramp up the value of $\alpha$ from 0 to $\alpha_{max}$. We also experiment with $K=\{3, 5\}$ for filtering the valid relation samples during the filtering process after each training iteration. The performance of the self-ensemble model does not vary much with these choices of $E$ or $K$. So we use $E=5$ and $K=3$ for final experiments.

\subsection{Results}

We include the results of our experiments in Table \ref{tab:results}. We run the CNN, PCNN, EA, and BGWA models 5 times with different random seeds and report the average with standard deviation in the `Student' column in Table \ref{tab:results}. The column `SEF' (Self-Ensemble Filtering) is the average results of 5 runs of CNN, PCNN, EA, and BGWA models with the self-ensemble filtering framework. We see that our SEF framework achieves $2.1 \%$, $1.1 \%$, $2.1 \%$, and $2.3 \%$ higher F1 score for the CNN, PCNN, EA, and BGWA models respectively compared to the Student models. If we compare the precision and recall score of the four models, we see that our self-ensemble framework improves the recall score more than the corresponding precision score in each of these four models. These results show the effectiveness of our self-ensemble filtering framework in a distant supervised dataset. 

\subsection{Self-Ensemble without Filtering}

We experiment with how the self-ensemble version of the student models behave without filtering the noisy samples after each iteration. So in this setting, we use the entire distant supervised training data at every iteration. The results are included in Table \ref{tab:ablation} under the `SE' (Self-Ensemble) column. This result shows that the performance of the four neural models under self-ensemble training without filtering is not much different from the `Student' performance of Table \ref{tab:results}. This shows that the filtering of the noisy samples from the training dataset helps to improve the performance of our proposed self-ensemble framework.

\begin{table}[ht]
\small
\centering
\begin{tabular}{l|cccc}
\hline
      & \multicolumn{4}{c}{SE}                                                                                                                                                                                      \\ \hline
Model & Prec.                                                         & Rec.                                                          & F1                                                            & $\downarrow$ \\ \hline
CNN   & \begin{tabular}[c]{@{}c@{}}0.448\\ $\pm$\\ 0.012\end{tabular} & \begin{tabular}[c]{@{}c@{}}0.610\\ $\pm$\\ 0.029\end{tabular} & \begin{tabular}[c]{@{}c@{}}0.516\\ $\pm$\\ 0.015\end{tabular} & 2.3\%        \\ \hline
PCNN  & \begin{tabular}[c]{@{}c@{}}0.432\\ $\pm$\\ 0.005\end{tabular} & \begin{tabular}[c]{@{}c@{}}0.670\\ $\pm$\\ 0.012\end{tabular} & \begin{tabular}[c]{@{}c@{}}0.525\\ $\pm$\\ 0.005\end{tabular} & 1.2\%        \\ \hline
EA    & \begin{tabular}[c]{@{}c@{}}0.421\\ $\pm$\\ 0.014\end{tabular} & \begin{tabular}[c]{@{}c@{}}0.647\\ $\pm$\\ 0.017\end{tabular} & \begin{tabular}[c]{@{}c@{}}0.510\\ $\pm$\\ 0.011\end{tabular} & 3.4\%        \\ \hline
BGWA  & \begin{tabular}[c]{@{}c@{}}0.424\\ $\pm$\\ 0.021\end{tabular} & \begin{tabular}[c]{@{}c@{}}0.689\\ $\pm$\\ 0.020\end{tabular} & \begin{tabular}[c]{@{}c@{}}0.524\\ $\pm$\\ 0.010\end{tabular} & 1.4\%        \\ \hline
\end{tabular}
\caption{Precision, Recall, and F1 score of the self-ensemble version of the student models on NYT dataset without noise filtering. We report the average of five runs with standard deviation. $\downarrow$ column shows the absolute \% decline of F1 score respect to the SEF models (Table \ref{tab:results}).}
\label{tab:ablation}
\end{table}

\begin{table}[ht]
\small
\centering
\begin{tabular}{l|cccc}
\hline
      & \multicolumn{4}{c}{Ensemble} \\ \hline
Model & Prec.    & Rec.     & F1  & $\downarrow$ \\ \hline
CNN   & 0.456    & 0.613    & 0.523 &  1.6\% \\ 
PCNN   & 0.437   & 0.679   & 0.532 &  0.5\%  \\
EA    & 0.454    & 0.658    & 0.537 &  0.7\% \\ 
BGWA  & 0.410    & 0.679    & 0.512 &  2.6\% \\ \hline
\end{tabular}
\caption{Precision, Recall, and F1 score of the ensemble version of the student models on NYT dataset. $\downarrow$ column shows the absolute \% decline of F1 score respect to the SEF models (Table \ref{tab:results}).}
\label{tab:ensemble_vs_selfensemble}
\end{table}

\subsection{Ensemble vs Self-Ensemble Filtering}

Since our SEF framework has an ensemble component, we compare its performance with the ensemble versions of the independent student models. The `Ensemble' column in Table \ref{tab:ensemble_vs_selfensemble} refers to the ensemble results of the 5 runs of each student model. We use the five runs of the models on the test data and average the softmax output of these runs to decide the relation. We see that our SEF framework outperforms the ensemble results for CNN, PCNN, EA, and BGWA with $1.6\%$, $0.5\%$, $0.7\%$ and $2.6\%$ F1 score respectively. Here, we should consider the fact that to build an ensemble model, the student models must be run multiple times (5 times in our case). In contrast, self-ensemble models can be built in a single run with little cost of maintaining the moving average of the student model. 

\section{Related Work}

There are two approaches for relation extraction \cite{Nayak2021DeepNA}: (i) Pipeline approaches \cite{zeng2014relation,zeng2015distant,jat2018attention,nayak2019effective} (ii) Joint extraction approaches \cite{takanobu2019hrlre,nayak2019ptrnetdecoding}. Most of these models work with distantly supervised noisy datasets. Thus noise mitigation is an important dimension in this area of research. Multi-instance relation extraction is one of the popular methods for noise mitigation. \newcite{riedel2010modeling}, \newcite{hoffmann2011knowledge}, \newcite{mimlre}, \newcite{lin2016neural}, \newcite{yaghoobzadeh2017noise}, \newcite{vashishth2018reside}, \newcite{wu2018improving}, and \newcite{ye2019intra} used this multi-instance learning concept in their proposed relation extraction models. For each entity pair, they used all the sentences that contain these two entities to find the relation between them. Their goal was to reduce the effect of noisy samples using this multi-instance setting. They used different types of sentence selection mechanisms to give importance to the sentences that contain relation specific keywords and ignore the noisy sentences. But this idea may not be effective if there is only one sentence for an entity pair. \newcite{Ren2017CoTypeJE} and \newcite{yaghoobzadeh2017noise} used the multi-task learning approach for mitigating the influence of the noisy samples. They used fine-grained entity typing as an additional task in their model.

\newcite{wu2017adversarial} used an adversarial training approach for the same purpose. They add noise to the word embeddings to make the model more robust for distantly supervised training. \newcite{Qin2018DSGANGA} used the generative adversarial network (GAN) to address this issue of the noisy samples in relation extraction. They used a separate binary classifier as a generator in their model for each positive relation class to identify the true positives for that relation and filter out the noisy ones. \newcite{qin2018robust} used reinforcement learning to identify the noisy samples for the positive relation classes. \newcite{He2020ImprovingNR} used reinforcement learning to identify the noisy samples for the positive relations and then use the identified noisy samples as unlabelled data in their model. \newcite{Shang2020AreNS} used a clustering approach to identify the noisy samples. They assign the correct relation label to these noisy samples and use them as additional training data in their model. Different from these approaches, we propose a student-teacher framework that can work with any supervised neural network models to address the issue of noisy samples in distantly-supervised datasets.

\section{Conclusion}

In this work, we propose a self-ensemble based noisy samples filtering framework for distantly supervised relation extraction. Our framework identifies the noisy samples during training and removes them from the training data in the following iterations. This framework can be used with any supervised relation extraction models. We run experiments using several state-of-the-art neural models with this proposed filtering framework on the distantly supervised New York Times dataset. The results show that our proposed framework improves the robustness of these models and increases their F1 score on the relation extraction task.

\section*{Acknowledgments}
This research is supported by A*STAR under its RIE 2020 Advanced Manufacturing and Engineering (AME) 
programmatic grant RGAST2003, (award \# A19E2b0098, project K-EMERGE: Knowledge Extraction, Modelling, and Explainable Reasoning for General Expertise).


\bibliographystyle{acl_natbib}
\bibliography{ranlp2021}


\end{document}